\documentclass[pmlr]{jmlr}% new name PMLR (Proceedings of Machine Learning)

\RequirePackage{graphicx}
 \usepackage{booktabs}
\makeatletter
\def\set@curr@file#1{\def\@curr@file{#1}} %temp workaround for 2019 latex release
\makeatother
\usepackage[load-configurations=version-1]{siunitx} % newer version

\theorembodyfont{\upshape}
\theoremheaderfont{\scshape}
\theorempostheader{:}
\theoremsep{\newline}

\jmlrvolume{[VOLUME \# TBD]}
\jmlryear{2024}
\newcommand{\bE}{\mathbb{E}}

\newcommand{\PropAlg}{\textsc{NestedRecommendation}}
\usepackage{makecell}
\usepackage{caption}
\usepackage{subcaption}
\usepackage{verbatim}
\usepackage{multirow}
\usepackage{array}
\usepackage{colortbl}
\usepackage{nicematrix}
\usepackage{float}
\usepackage{enumitem}
\usepackage{placeins}
\emergencystretch=\maxdimen
\title[Gender Fairness in ML-driven Chronic Pain Care]{Investigating Gender Fairness in Machine Learning-driven Personalized Care for Chronic Pain}

\author{%Anonymous authors
\Name{Pratik Gajane}
       \Email{p.gajane@tue.nl}\\ 
       \addr Department of Mathematics and Computer Science,\\
  Eindhoven University of Technology,\\
  Eindhoven, the Netherlands.
\AND
  \Name{Sean Newman}
\Email{scnewman@umich.edu}\\
\addr School of Public Health, \\ 
  University of Michigan, \\
Ann Arbor, Michigan, USA.
\AND
\Name{Mykola Pechenizkiy}
\Email{m.pechenizkiy@tue.nl}\\ 
\addr Department of Mathematics and Computer Science,\\
Eindhoven University of Technology,\\
Eindhoven, the Netherlands.
      \AND
       \Name{John D. Piette}
       \Email{jpiette@umich.edu}\\ 
       \addr Department of Health Behavior and Health Equity,\\
       School of Public Health, University of Michigan, \\
       Ann Arbor, Michigan, USA.
} 
\begin{document}
\maketitle
\begin{abstract}
Chronic pain significantly diminishes the quality of life for millions worldwide. 
While psychoeducation and therapy can improve pain outcomes, many individuals experiencing pain lack access to evidence-based treatments or fail to complete the necessary number of sessions to achieve benefit. 
Reinforcement learning (RL) shows potential in tailoring personalized pain management interventions according to patients’ individual needs while ensuring the efficient use of scarce clinical resources. However, clinicians, patients, and healthcare decision-makers are concerned that RL solutions could exacerbate disparities associated with patient characteristics like race or gender. In this article, we study \textit{gender fairness} in personalized pain care recommendations using a real-world application of reinforcement learning \citep{Piette2022}. Here, adhering to gender fairness translates to minimal or no disparity in the utility received by subpopulations as defined by gender. We investigate whether the selection of relevant patient information (referred to as features) used to assist decision-making affects gender fairness. Our experiments, conducted using real-world data \citep{piette2022dataset2}, indicate that included features can impact gender fairness. Moreover, we propose an RL solution, \PropAlg, that demonstrates the ability\thinspace: i) to adaptively learn to select the features that optimize for utility and fairness, and ii) to accelerate feature selection and in turn, improve pain care recommendations from early on, by leveraging clinicians' domain expertise.
\end{abstract}

\section{Introduction}
\label{sec:Int}
%Pain is an unpleasant sensation and emotional experience that leads to poor quality of life for millions of people worldwide. Chronic pain is a prevalent and increasingly common problem that is unequally distributed in society based on many factors including gender \citep{Andrews2017, 00006396-202209000-00013}.  Chronic pain, as indicated by a World Health Organization study, is associated with a four-fold increase in the occurrence of depression or anxiety \citep{Slack2022}. Chronic pain frequently disrupts individuals' capacity to function in key life roles such as work and family, engage in pleasurable activities that give life meaning, and maintain a healthy lifestyle. %particularly in the areas of sleep hygiene and physical activity. 

Chronic pain is a widespread issue impacting millions of individuals worldwide and its distribution within society is markedly unequal, often influenced by demographic factors, including gender \citep{Andrews2017, 00006396-202209000-00013}.
Cognitive behavioral therapy (CBT) and related approaches are the most common, evidence-based non-pharmacologic treatments for chronic pain \citep{Chou2008}. 
%Through CBT, pain patients learn strategies to manage stressful thoughts (e.g., anxiety about pain's impact on their relationships) and adapt behaviors that decrease pain's impact on their quality of life and functioning. 
Pain CBT is typically delivered by a trained therapist to individuals and groups, either through face-to-face sessions or using remote technology. Due to the necessity of multiple (weekly) sessions and a limited availability of therapists, many patients suffering from chronic pain lack access to pain CBT services or discontinue treatment before experiencing therapeutic benefits.

Considering the significant impact of chronic pain on individuals and society, and the challenges of providing effective treatment to the large number of patients who could benefit, a variety of machine learning (ML) solutions have been proposed to improve the quality and scalability of pain psychotherapies. 
%See \cite{Matsangidou2021} for a comprehensive survey. 
Machine learning could be especially beneficial in optimizing the delivery of pain CBT  to better address the considerable variability among patients, encompassing their response to treatments and susceptibility to adverse effects \citep{Ablin2012}. This variability limits the efficacy of generic interventions, emphasizing the necessity for more personalized approaches to address the unique needs, preferences, and responsiveness of each patient with chronic pain.  

Thus by its nature, chronic pain care necessitates personalized treatments over an extended duration taking the feedback from patients into consideration. Reinforcement learning (RL) algorithms are especially suited to make such sequential decisions using feedback from patients, including passively-collected or patient-reported information about their behaviors, treatment engagement, and health status. RL also has shown the ability to personalize pain care recommendations \citep{Komorowski2018, Saria2018, ROGGEVEEN2021102003}. 
%
%For example, \cite{Komorowski2018} used reinforcement learning solutions to recommend personalized treatments that are on average more reliable than human clinicians. \cite{Saria2018} proposed reinforcement learning methods for individualized treatment strategies to correct hypotension in Sepsis. \cite{ROGGEVEEN2021102003} propose similar solutions for optimizing hemodynamic treatment for critically ill patients with Sepsis.  
%
In this article, we draw upon the study of \cite{Piette2022} who conducted a randomized trial to evaluate the effectiveness of an RL strategy to deliver personalized  CBT for chronic pain. 
This approach automatically adjusts the type and duration of treatment sessions based on patients’ feedback regarding their pedometer-measured step counts and pain-related interference. They found that the RL algorithm was effectively able to personalize the intensity of sessions to participants and that on average patient-reported outcomes improved as the program gained experience through interactions \citep{PIETTE2022100064}. 
Compared to standard telephone CBT delivered by a therapist, the RL-supported CBT resulted in non-inferior improvements in pain-related disability at three months (primary outcome), while using less than half the therapist time \citep{PIETTE2022100064}. At six months, significantly more patients randomized to the RL solution had clinically meaningful improvements in pain-related disability than patients receiving standard care ($37\%$ versus $19\%$, $p<.01$) \citep{Piette2022}. Also, a greater proportion of patients receiving RL-supported CBT had clinically meaningful 6-month improvements in pain intensity ($29\%$ versus $17\%, p=.03$).

While these results are encouraging,  a possible drawback of RL-supported healthcare services is that they may introduce/amplify biases that limit access to care or create suboptimal outcomes for certain patient demographics. In general, the design of the majority of precision healthcare ML algorithms ignores the sex and gender dimensions, and their contribution to health differences among individuals \citep{Cirillo2020}. In the context of the study of \cite{Piette2022}, conducted within the United States Department of Veterans Affairs healthcare system, women represent a relatively small subgroup of patients, and their unique experiences regarding pain care might affect their engagement and utility gained from CBT pain interventions \citep{KehleForbes2017}. This might potentially affect gender fairness in the utility gained from CBT interventions. 

To see how we assess gender fairness in this study, note that, posed as an RL problem, the utility of a pain care recommendation to a patient is represented by a numerical value called ``reward" (a detailed formulation will be given in Section \ref{sec:Formalization}).  
%The higher the reward, the higher the utility of the associated recommendation to the patient. 
In this study, we consider adhering to gender fairness as minimal or no disparity between the average rewards received by subpopulations as defined by gender. Other definitions of fairness have been considered in the literature \citep{Grote2022}. 

Gender biases in treatment targeting and personalization could arise from several problems in the development and use of RL algorithms to personalize healthcare. For example, significant differences between a training sample and the population of patients for which the algorithm will be used could introduce unfair or undesirable biases in RL decisions. Differential feedback from participating patients in which under-served and priority subgroups fail to provide input at the same level of accuracy and timeliness as other patients also could introduce biases. Finally, substantial concerns arise from limitations in the design of RL algorithms, particularly regarding the inclusion and operationalization of features (i.e., relevant patient information) in decision-making.

%JP-In this paper we may only want to focus on the final issue mentioned above (selection of state features). However two other issues could be mentioned and our conversations have made me think about them...let's bookmark these for future conversations and perhaps papers/analyses: (1) the impact of the magnitude of alpha (exploration) as you mention below. I think this is interesting and I hadn't thought of it before. (2) the magnitude of the "cost" function used to penalize more human-resource action choices.  That cost coefficient (I know there are a few ways to introduce cost into the equation) was fairly arbitrary in our study. In the extreme the cost function would totally drive the action choice possibly to the disadvantage of women or other groups.  For example, if women are far more likely to benefit from longer human interactions, but we assign those human interactions an infinite cost, then the pattern of care could ignore womens' needs.

 \subsection{Our Contributions}
 \label{sec:OurCont}
 In this study, we investigate gender fairness in personalized pain care recommendations made using a reinforcement learning algorithm.  In particular, we examine the extent to which the third potential pitfall of RL-supported healthcare services noted above - namely suboptimal specification of features - could contribute to biased decision-making. We utilize the dataset \cite{piette2022dataset2} which is also used in an extensive study on RL-driven pain care \citep{PIETTE2022100064}. To allow for personalized recommendations, we model the problem at hand using \textit{contextual bandits}, a reinforcement learning problem formulation well-suited to deal with personalized sequential decision-making. In this model, the algorithm leverages a range of patient-specific features to tailor its recommendations. 
%Our results indicate that although adjustments to the parameter utilized by LinUCB result in changes to the outcomes of pain care recommendations produced by the algorithm, this impact is consistent across both men and women.  However, 
 %Our results indicate that when the algorithm lacks access to certain patient information, notably self-reported pain measurements, the quality of pain care recommendations for women is discernibly different compared to men. 
 Our results indicate that when the RL algorithm makes decisions based on a certain set of features, the utility of pain care recommendations for women is discernibly different compared to men \footnote{Please note that the results provided are based on the genders recorded in the dataset \cite{piette2022dataset2}\thinspace: men and women.}.
Furthermore, we propose a generalizable, data-driven reinforcement learning algorithm to optimize feature selection and mitigate biases in utility across genders. 
%Our algorithm has the ability to expedite feature selection and in turn, improve pain care recommendations by leveraging clinicians' domain expertise. ``Cold start" here refers to the phenomenon of subpar initial performance for reinforcement learning solutions. 
Our algorithm has the ability to expedite feature selection and in turn, improve pain care recommendations from the outset by leveraging clinicians' domain expertise. This approach helps mitigate the problem of initial subpar performance, known as the ``cold start", that often hinders the real-world applicability of RL solutions.
While the focus of this article is on gender, our algorithm can also be applied to address biases across other relevant factors. 

% This work is a joint effort between machine learning researchers and healthcare researchers. Accordingly, we endeavored to ensure that both the problem formulation and our approach remain rooted in the healthcare domain while also being feasible from a machine learning standpoint. 
 
\subsection{Generalizable Insights about Machine Learning in the Context of Healthcare}
\label{sec:GenInsights}
%{\color{red} %-This section is \emph{required}, must keep the above title, and should be the final part of your introduction.  In about one paragraph, or 2-4 bullet points, explain what we should \emph{learn} from readingthis paper that might be relevant to other machine learning in healthendeavors.} For example, a work that simply applies a bunch of existing algorithms to a new domain may be useful clinically but doesn't increase our understanding of the machine learning and healthcare; if that study also investigates \emph{why} different approaches have different performance, that might get us excited!  A more theoretical machine learning work may be in how it enables a new kind of clinical study. Reviewers and readers will look to evaluate (a) the significance of your claimed insights and (b) evidence you provide later in the  work of you achieving that contribution}

Previous research (including studies cited in Section \ref{sec:RelatedWork} below) has demonstrated that machine learning, particularly reinforcement learning, has the potential to enhance the efficiency of complex healthcare services like CBT. This includes limiting the delivery of marginal-impact health services while maximizing the utilization of available clinical resources to benefit a larger number of patients. However, serious concerns have been raised about the impact of these algorithms on disparities in treatment access, quality, and outcomes affecting socioeconomically vulnerable groups of patients \citep{McCradden2020, Parikh2019,Daneshjou2021, Lyles2023}. The extent to which such disparities exist, or even how they should be identified is not well understood and almost no work has offered generalizable solutions to addressing fairness-related problems in RL-based healthcare solutions.

In this article, we empirically demonstrate disparities across genders in the performance of an RL algorithm when using various subsets of relevant patient information. Here, performance is measured both in terms of the average reward and suboptimal pain care recommendations.  While gender-based bias in health outcomes is an important concern, our focus on gender fairness represents a much broader set of social determinants of health that could reflect unacceptable biases in health systems. % goals of equal access and outcomes across patient groups.

Additionally, we propose a generalizable algorithmic RL solution that can be applied in conjunction with expert inputs to optimize population-level outcomes while minimizing disparities in outcomes across subgroups. Our algorithm can make use of valuable inputs from healthcare experts upfront, enabling improved pain care recommendations from early on. This helps to address the issue of low initial performance of RL algorithms which may limit their practical effectiveness in real-world pain care management. In this context, low initial performance would translate to the possibility of initial patient interactions receiving poor recommendations, potentially resulting in inadequate pain care and diminished patient outcomes. Hence, approaches such as ours, which seek to enhance the initial performance of RL-based healthcare solutions, are pivotal for refining treatment strategies and enhancing patient satisfaction and well-being. Moreover, our approach would bolster the confidence of healthcare professionals in RL-based solutions, as it empowers them with a degree of control over the algorithmic process to rectify its initial erroneous decisions. 
%that might make incorrect decisions especially based on limited or biased input data
%
%RL algorithms encounter subpar initial performance due to exploration, which refers to their tendency to try out various actions to learn their effectiveness. However, this initial exploration phase often leads to mistakes, resulting in suboptimal action selection and subsequent underperformance. By leveraging clinicians' domain expertise, we provide the algorithm with valuable insights from healthcare professionals upfront, enabling improved pain care recommendations right from the start. %In short, our algorithm has the ability to improve pain care recommendations from the outset by utilizing the expertise of clinicians in the field

By documenting potential problems of gender fairness related to feature selection along with some potentially useful solutions, we hope to inform the broader debate regarding fairness-related consequences of machine learning in healthcare resource allocation and patient management (in the current application, fairness or equity across gender groups).  We hope that these strategies can facilitate the effective application of RL solutions in healthcare problems affecting millions of patients (here, pain patients receiving pain psychotherapy) while also assuring clinicians and decision-makers that such solutions can, in fact, reflect community values such as gender-based equity.

\section{Related Work}
\label{sec:RelatedWork}
%-Make sure you also put your work in the context of related work.  Who else has worked on this problem, and how did they approachit?  What makes your direction interesting or distinct?
Machine learning has been found useful in pain-related assessment, prognosis, and self-management support \citep{Zhang2023, Singhal2023, Zakeri2022}. Some of the most advanced work has focused on algorithms designed to risk-stratify patients and identify unexplained variations in outcomes across subgroups. For example, deep learning algorithms have been found to be useful in explaining socioeconomic disparities in pain severity by incorporating predictors that go beyond standard clinical assessments such as radiography \citep{Pierson2021}.
On the other hand, \cite{doi:10.1126/science.aax2342} illustrate that commercially available prediction models may exacerbate racial disparities by using cost as a biased proxy for clinical severity.

Far fewer studies represent actual clinical trials in which ML solutions have been evaluated as a component of patient care. A recent systematic review identified 152 out of 627 trials of AI applications in healthcare incorporated some form of patient-reported outcome, with the majority of those studies published since 2021 \citep{Pearce2023}. Some evidence suggests that patient messaging based on reinforcement learning can improve self-care behaviors (e.g., encouraging physical activity among patients with diabetes) relative to a fixed algorithm \citep{YomTov2017,Hochberg2016}. The use of reinforcement learning for physical activity encouragement in cardiac rehabilitation is explored by \cite{10.1007/978-3-031-42286-7_11}. 
\cite{634483} proposed a deep reinforcement learning approach for personalized opioid dosing recommendations for pain care. 
Despite the growing interest in the field, there exists a paucity of studies documenting real-world impact on treatment quality, health behaviors, or clinical outcomes to generalize about the effectiveness of these applications \citep{Han2023}.

Along with the enthusiasm about the potential of machine learning to improve healthcare, there has emerged a growing mistrust, skepticism, or even fear about the loss of human control of healthcare resource allocation, patient communication, and clinical decisions \citep{McCradden2020, Parikh2019,Daneshjou2021, Lyles2023}. In particular, \cite{Tewari2017} point out a number of challenges that should be addressed in order to maximize the usefulness of contextual bandits (and other machine learning approaches) in the healthcare domain, including generalizable methods for assessing the suitability of included features.
In a 2019 review, \cite{Triantafyllidis2019} identified only 3 randomized trials in which ML approaches were evaluated as part of a digital health intervention. Two of the three were given a global quality rating of ``weak." 
%Given the very limited number of empirical analyses of the fairness problem in healthcare using real-world samples, the legitimate concerns among clinicians and policymakers too often lead to an unproductive sense of inertia regarding the exploration or adoption of machine learning-based programs that may dramatically increase the quality, accessibility, or efficiency of health services. 
%PG(8 April): John, I wonder if we could quickly mention some of the articles about fairness concerns in healthcare and discuss how our article is distinct from theirs? https://www.nature.com/articles/s41591-020-01192-7 seems particularly relevant. [JP: I  NOWHIGHLIGHTED THIS STUDY ABOVE UNDER RELATED WORK...THE LANGUAGE THEY USE IS A LITTLE MISLEADING...THEY TALK ABOUT "REDUCING UNEXPLAINED PAIN DISPARITIES"...BUT WHAT THEY ACTUALLY MEAN IS REDUCTING THE UNEXPLAINED STATISTICAL VARIATION IN PAIN ACROSS GROUPS]
%
%PG-JP RESPONSE We might also want to mention -- Obermeyer et al.``Dissecting racial bias in an algorithm used to manage the health of populations", because it seems very influential (over 3500 citaions!) [YES IMPORTANT STUDY BUT WE'RE MAKING AN IMPORTANT DISTINCTION HERE...OBERMEYER AND THE NATURE STUDY ABOVE ARE ABOUT PREDICTION RATHER THAN USING RL/MACHINE LEARNING AS PART OF AN ACTUAL INTERVENTION. I'VE NOW HIGHLIGHTED THIS OBERMEYER STUDY ABOVE UNDER RELATED WORK]
Given the scarcity of empirical analyses of fairness concerns in ML-based healthcare using real-world samples, legitimate concerns among clinicians and policymakers arise. These concerns often lead to an unproductive sense of inertia regarding the exploration or adoption of ML-based programs that may greatly increase the quality, accessibility, or efficiency of health services.

%Todo for PG -- differentiate from https://arxiv.org/pdf/2101.09460.pdf 
Our proposed method that learns to select the features that optimize for fairness and utility is similar in objective to previous work on RL-based feature selection \citep{HAZRATIFARD20131892, rasoul2021feature, 9507351}. However, our proposed method can use domain expertise to accelerate feature selection, unlike these solutions. For fairness concerns in the general RL literature and RL-based healthcare services, see \cite{gajane2022survey} and \cite{10.1145/3609502} respectively.

\section{Formalizing Pain Care as a Reinforcement Learning Problem}
\label{sec:Formalization}
We model pain care using \textit{contextual bandits} \textemdash\ a reinforcement learning formalization that has been used toward the goal of personalized decision-making in many domains including healthcare. In a contextual bandits problem, at each decision step $t=1,2,\dots, T$\thinspace: 
\begin{itemize}
    \item The algorithm observes a context i.e., some relevant information in the form of a set of feature values.
    \item Using this contextual information and previously observed feedback, the algorithm selects an option $a_t$ from the set of available options. After selecting an option, the algorithm receives a (possibly randomized) numerical value $r_{t,a_t}$ as feedback. In reinforcement learning terminology, options and feedback are called \textit{actions} and \textit{rewards} respectively. %The reward for each action is typically assumed to be stochastic.  
    It should be noted that the received reward $r_{t,a_t}$ only corresponds to the action taken $a_t$ and no reward is observed at time $t$ for actions $a\neq a_t$. Reward $r_{t,a_t}$ informs the algorithm about the quality of $a_t$. The higher the received reward, the better the corresponding action. 
    \item The algorithm may choose to improve its action selection strategy with the new observation\thinspace: \{observed context, selected action, received reward\}. 
\end{itemize}

\begin{comment}  
With this formulation, the algorithm's goal can be expressed as maximizing the cumulative sum of received rewards $\sum_{t=1}^{T}r_{t, a_t}$. Alternatively, the algorithm's goal can also be expressed as minimizing \textit{regret}
formally defined by
\begin{equation*}
    \bE\left[ \sum_{t=1}^{T} r_{t, a^*_t} \right] - \bE\left[ \sum_{t=1}^{T} r_{t, a_t} \right],
\end{equation*}
where $\bE$ denotes the expectation and $a^*_t$ is the action with the maximum expected reward at decision step $t$.
Regret can be understood as the total mistake cost. %That is, at every decision step $t$ at which the algorithm selects an action $a_t \neq a^*_t$, the algorithm incurs some regret. 
\end{comment}

For personalized pain care, a decision step corresponds to the beginning of a patient interaction. Context/features can be understood as the relevant patient information. Actions are the available pain care recommendations and feedback corresponds to the effects of the given recommendation on the patient. 

For our study, we used the dataset provided in \cite{piette2022dataset2}. 
For each patient interaction, the following patient information is given in the dataset\thinspace:
\begin{center}
    \begin{tabular}{l  l }
    $\%$ days this week with steps goal met & 
    CBT skill practice this week \\ 
    Sleep quality this week & 
    Sleep duration this week \\
    Pain interfere1 & 
    Pain interfere2 \\
    Pain intensity change  &
    Session Number 
    \end{tabular}
\end{center} 

\begin{comment}
    \begin{itemize}
    \item Percentage of days in the current week with steps goal met %(0-1)  
    \item Pain intensity change %(0-1) 
    \item CBT skill practice this week %(0-10) 
    \item Sleep quality this week %(0-10) 
    \item Sleep duration this week %(0-22) 
    \item Pain interfere1 %(0-10) 
    \item Pain interfere2 %(0-10) 
    \item Session Number %(1-10) 
\end{itemize}    
\end{comment}

%\todo{Information about the features}
%In the above, Pain interfere1 represents a self-reported response to the query -- ``What number best describes how much pain has interfered with your enjoyment of life today, with 0 meaning pain does not interfere and 10 meaning pain completely interferes?”. Moreover, Pain interfere2 is a self-reported response to the query -- ``What number best describes how much pain has interfered with your general activity today, with 0 meaning pain does not interfere and 10 meaning pain completely interferes?”

These features were decided upon in consultation with healthcare experts. 
%In the above, Pain interfere1 represents a self-reported response to the query -- ``What number best describes how much pain has interfered with your enjoyment of life today, with 0 meaning pain does not interfere and 10 meaning pain completely interferes?”. Moreover, Pain interfere2 is a self-reported response to the query -- ``What number best describes how much pain has interfered with your general activity today, with 0 meaning pain does not interfere and 10 meaning pain completely interferes?”. 
More details about the features and how they are collected can be found in Appendix \ref{app:Features}. 

The recommendation options (or actions, in the parlance of reinforcement learning) are\thinspace:
\begin{itemize}
    \item Option 1\thinspace: Interactive voice response (IVR) call. During IVR calls, patients hear a recorded message from their therapist.
    \item Option 2\thinspace: A 15-minute telephone session with a therapist.
    \item  Option 3\thinspace: A 45-minute telephone session with a therapist. 
\end{itemize}
%It is apparent that option 2 is more human-resource intensive than option 1, while option 3 is the most human-resource intensive among them. 
%To address this disparity, \cite{PIETTE2022100064} recommend discounting the expected rewards of option 2 and option 3 with additive factors of $-0.02$ and $-0.06$ respectively. 

%PG see my comment above about finding a more empirically-derived, defensible, and fairness-promoting cost coefficient.

For each patient interaction, the entry recorded in the dataset contains values of all the features mentioned above, the recommended action, and the observed reward. 
%The rewards mentioned in the dataset are computed \todo{How are the rewards in the dataset computed?}
%For SN: Sean, Could you please provide a brief explanation of how the rewards recorded in the provided dataset were calculated? 
The rewards were computed by adding the change in pedometer steps (scaled between 0-0.5) to the average of the two pain interference questions (scaled between 0-0.5). 
%Finally, a small penaly was subtracted per minute of clinician time (whether 15 or 45) used.  
In the above, option 2 is more human-resource intensive than option 1, while option 3 is most human-resource intensive. 
To mitigate this disparity, \cite{Piette2022} recommend discounting the rewards of option 2 and option 3 with additive factors of $-0.02$ and $-0.06$ respectively. 
\section{Methods}
%\textcolor{lightgray}{Tell us your techniques!  If your paper is develops a novel machine learning method or extension, then be sure to give the technical details---as you would for a machine learning publication---here and, as needed, in appendices.  If your paper is developing new methods and/or theory, this section might be several pages. If you are combining existing methods, feel free to cite other packages and papers and tell us how you put them together; that said, the work should stand alone for someone in that general machine learning area.  }

%\textcolor{lightgray}{\emph{Lack of technical details, such that the soundness of the methods can be verified, is a major reason that otherwise strong-looking papers are scored low/rejected.}}

In this section, we provide details of the methods we used in our study. 
\subsection{Use of Existing Method}
The purpose of our first set of experiments was to verify the influence of the set of features included in the context on the performance of the algorithm in terms of its gender fairness and utility. Here we used the algorithm LinUCB \citep{LinUCB}. We include it in Appendix \ref{app:LinUCB} for completeness.

%LinUCB operates by maintaining a set of linear models, each corresponding to an action, and updates these models based on observed features and associated rewards. LinUCB employs an upper confidence bound strategy to balance exploration and exploitation, wherein it chooses actions with higher uncertainty within the bounds of linear model predictions, thus exploring potentially advantageous actions while leveraging existing knowledge. By iteratively updating model parameters and selecting actions according to their estimated values and uncertainty, LinUCB aims to maximize cumulative rewards over time. 
LinUCB maintains linear models for actions, adjusting them with observed features and rewards. It selects actions with uncertainty within model bounds to maximize cumulative rewards through iterative updates based on estimated values.
We chose this algorithm as it is highly influential and forms the basis of other RL methods. Accordingly, we expect that the results obtained using LinUCB will have wide-reaching pertinence and will also be relevant while using other solutions based on LinUCB. 
%Using LinUCB, we show that the set of features included in the context can influence the performance of the algorithm in terms of its gender fairness and utility.  

\subsection{Our Proposed Method}
In this section, we describe our proposed solution, given in Algorithm \ref{alg:PropAlg}, to dynamically select the optimal feature set from the provided options. The identity of the optimal feature set is determined by an arbitrary performance criterion depending on received rewards. 
\begin{algorithm}
    \caption{\PropAlg}\label{alg:PropAlg}
    \label{alg:Nested}
    \SetAlgoLined
    \SetKwInOut{Input}{Input}
    \Input{Feature sets, pain care recommendations, T, Policy1, Policy2, performance criteria.}
    \For{$t = 1$ to $T$}{
        \tcp*[l]{\underline{Level 1: Selection of Feature Sets}}
        Use action selection step of Policy1 to choose $\text{set}(t)$ from the given feature sets.\;
        
        \tcp*[l]{\underline{Level 2: Selection of Pain Care Recommendations}}
        Use Policy2 with $\text{set}(t)$ to choose the recommendation $\text{rec}(t)$ from the given pain care recommendations.\;\\
        Apply $\text{rec}(t)$ and note the received reward $r(t)$.\; \\
        Use $\text{rec}(t)$ and $r(t)$ to update the internal parameters of Policy2 (to be used in future selections at Level 2).\; \\
        Use $\text{set}(t)$ and $r(t)$ to perform the policy update step of Policy1 (to be used in future selections at Level 1).\;
    }
\end{algorithm}

In this nested strategy, the algorithm takes decisions at the following two levels.
\begin{itemize}
    \item At the first level\thinspace: The decision choices (or, actions) are the given feature sets, and the algorithm's objective is to learn to select the optimal feature set using history (i.e., previously selected feature sets and observed rewards).
    \item At the second level\thinspace: The decision choices are pain care recommendations, and the algorithm's objective is to learn to select the optimal pain care recommendation using history (i.e., previously selected recommendations and observed rewards). 
\end{itemize}

An instantiation of Algorithm \ref{alg:Nested} that allows us to make use of clinicians' domain knowledge about the feature sets is to use a Bayesian policy like Thompson Sampling (given in Appendix \ref{app:TS}) as Policy1 along with LinUCB as Policy2. In this instantiation, we can employ Beta distribution as a prior distribution to encode clinicians' domain knowledge about the suitability of feature sets as follows \footnote{For a short primer on Beta distribution and its use as a prior distribution, see \cite{BetaDist}.} \thinspace: For action $a$, use Beta$(\alpha_a, \beta_a)$ to encode the belief that the expected reward from action $a$ would be $\alpha_a/(\alpha_a + \beta_a)$.  Other ways to incorporate clinicians’ domain knowledge about the feature sets into prior distributions, as discussed in \cite{JOHNSON2010355}, can also be accommodated within \PropAlg. 
 %In the Thompson Sampling policy, at time step $t=1$, $\theta$ samples are drawn from the prior distribution indicating healthcare experts' beliefs about the actions (i.e., feature sets). 
 \begin{comment}
\begin{algorithm}
    \caption{Thompson Sampling} 
    \label{alg:TS}
    \SetAlgoLined
    \SetKwInOut{Input}{Input}
    \SetKwInOut{Output}{Output}
    \Input{Action set $\{1, 2, \dots, K\}$, parameters $\alpha_a$, $\beta_a$ for $a = 1, 2, \dots, K$}
    %\Output{Action to take $a(t)$}
    \BlankLine
    \For{time step $t=1,2, \dots$}{
         \underline{\textbf{Action Selection}}\; \\
        \For{each action $a = 1, 2, \dots, K$}{
            Sample $\theta_a(t)$ from the Beta$(\alpha_a, \beta_a)$ distribution\;
        }
        Select action $a(t) = \arg\max \theta_a(t)$ and receive reward $r(t)$\; \\
        \BlankLine
        \underline{\textbf{Policy Update}}\; \\
        Perform a Bernoulli trial with success probability $r(t)$ and observe sample $s(t)$\; \\
        \If{$s(t) = 1$}{
            $\alpha_{a(t)} = \alpha_{a(t)} + 1$\;
        }
        \Else{
            $\beta_{a(t)} = \beta_{a(t)} + 1$\;
        }
    }
\end{algorithm}
\end{comment}

\section{Cohort Selection and Data Extraction}
%-{\color{orange} \emph{This section is optional, and more theoretical work may not need  this section.  However, if you are using health data, then you need  to describe it carefully so that the clinicians can validate the  soundness of your choices.} } Describe the cohort.  Give us the details of any inclusion/exclusion criteria, what data were extracted, how features were processed, etc.  Recommended headings include:

%\subsection{Cohort Selection} 
Details about the study design including patient eligibility, intervention delivery, and the underlying RL approach used by the intervention are described in the main outcomes report \citep{Piette2022}, protocol paper \citep{Piette2016}, and a paper describing secondary analyses of the RL system’s functioning \citep{PIETTE2022100064}. In brief, the data used in this article came from the RL-supported segment of a randomized trial designed to assess the effectiveness of RL-supported CBT compared to standard therapist-delivered telephone CBT for chronic back pain management. Conducted among 278 patients recruited from two United States Department of Veterans Affairs healthcare systems, the study offered all participants a 10-week program of pain CBT. In the RL-supported intervention segment, patients' feedback on their physical activity (pedometer-measured step counts) and pain-related interference was collected through daily automated calls and used to inform the RL model's weekly selection from among the possible options as given in Section \ref{sec:Formalization}. 
%The three possible options each week were\thinspace: therapist-delivered telephone CBT sessions of varying durations (45 minutes or 15 minutes), or personalized therapist messages delivered through an automated call. 
Patients in the comparison group received ten 45-minute therapist-delivered telephone CBT sessions as recommended by professional guidelines. Outcome assessments, including patient-reported measures, were conducted at 3- and 6-months post-randomization to evaluate intervention efficacy and patient satisfaction.

%\subsection{Data Extraction} 
\paragraph{Data Extraction.}
The data was collected via questions asked during daily IVR surveys with the responses for each person averaged across days within a week. The features described in Section \ref{sec:Formalization} and Appendix \ref{app:Features} correspond to questions in the daily survey. They were based on domain expert advice about which data the RL algorithm should consider. %in choosing among the 3 possible intervention strategies. 

%\subsection{Feature Choices} 

\section{Experimental Results} 
We conducted two sets of experiments. The aim of the first set of experiments was to determine whether the features included in the context information affect gender fairness. The second set of experiments aimed to determine whether our proposed method \PropAlg \ can dynamically select the optimal feature set, where optimality is determined by an arbitrary performance criterion combining utility and fairness.

\subsection{Preprocessing}
%Directly utilizing a logged dataset for our evaluation experiments presents the challenge that rewards are solely recorded for actions chosen by the logging policy, which may diverge from those selected by the algorithm under evaluation. To address this challenge, we devised a model to emulate patient interactions based on the interactions outlined in the dataset \cite{piette2022dataset2}.   
Details about data preprocessing and the patient interactions can be found in Appendix \ref{app:Pre}. We ensured that the distribution of the total population across genders remains consistent with the dataset \cite{piette2022dataset2}. Overall, women constitute $12.5\%$ of the total number of patients. 

\subsection{Algorithmic Parameter Used in LinUCB}
The only parameter used in LinUCB is the value $\alpha$ (see Appendix \ref{app:LinUCB}). 
We used $\alpha = 0.3$ in all the results reported in this article. The role of $\alpha$ in the algorithm and our parameter selection process is explained in Appendix \ref{app:LinUCBParam}. 

\subsection{Competing Feature Sets} 
\begin{table}
\caption{Feature sets and included features}
\centering
\begin{NiceTabular}{c c c}
\toprule
Set 1 & Set 2 & Set 3 \\
\midrule
                \makecell[l]{
                Pain intensity change, \\ 
                CBT skill practice this week,  \\
                Sleep quality this week, \\
                Sleep duration this week, \\
                Pain interfere1, \\
                Pain interfere2, \\ 
                \% of days in current week \dots \\ \dots with steps goal met, \\
                Session number
                }  & 
                \makecell[l]{
                Pain intensity change, \\ 
                CBT skill practice this week,  \\
                Sleep quality this week, \\
                Sleep duration this week, \\
                \% of days in current week \dots\\ \dots with steps goal met, \\
                Session number
               } &
                \makecell[l]{ 
                            Pain intensity change, \\ 
                            CBT skill practice this week,  \\
                            Sleep quality this week, \\
                            Sleep duration this week, \\
                            Pain interfere2, \\
                             \% of days in current week \dots \\ \dots with steps goal met, \\
                            Session number  
                }                 \\
                \hline \hline
Set 4 & Set 5 & Set 6 \\
\hline
                \makecell[l]{
                            Pain intensity change, \\ 
                            CBT skill practice this week,  \\
                            Sleep quality this week, \\
                            Sleep duration this week, \\
                            Pain interfere1, \\
                             \% of days in current week \dots \\ \dots with steps goal met, \\
                            Session number  
                } 
                & 
                \makecell[l]{
                Pain intensity change, \\ 
                CBT skill practice this week,  \\
                Pain interfere1, \\
                Pain interfere2, \\
                 \% of days in current week \dots \\ \dots with steps goal met, \\
                Session number}         
               &
                \makecell[l]{Pain intensity change, \\ 
                CBT skill practice this week,  \\
                Sleep quality this week, \\
                Sleep duration this week, \\
                Pain interfere1, \\
                Pain interfere2, \\
                Session number}
                \\       
 \bottomrule
 \end{NiceTabular}
 \label{tab:FeatureSets}
\end{table}

We compared the results across the various feature sets shown in Table \ref{tab:FeatureSets}. We also conducted experiments using other combinations of features, with each combination containing at least three features.   
%with three more feature sets -- \{Pain intensity change, Sleep quality this week, Sleep duration this week, Pain interfere1, Pain interfere2, Session Number\}, \{of days in current week with steps goal met, CBT skill practice this week, Session Number\}, and \{\% of days in current week with steps goal met, CBT skill practice this week, Sleep quality this week, Sleep duration this week, Pain interfere1, Pain interfere2, Session Number\}. 
The findings from the six feature sets shown in Table \ref{tab:FeatureSets} proved particularly insightful. Results for other feature sets are available from the authors on request.

\subsection{Results\thinspace: Effects of Features on Gender Fairness} 
In the first set of experiments, we see how the features included in the context information can affect gender fairness. %This is important because, as noted in Section \ref{sec:Int}, reinforcement learning solutions for healthcare (including health services for patients with chronic pain) have been shown to be susceptible to biases based on patient attributes associated with social determinants of health, including gender and race.
\begin{figure}
\begin{minipage}{0.497\textwidth}
    \includegraphics[width=1.1\linewidth]{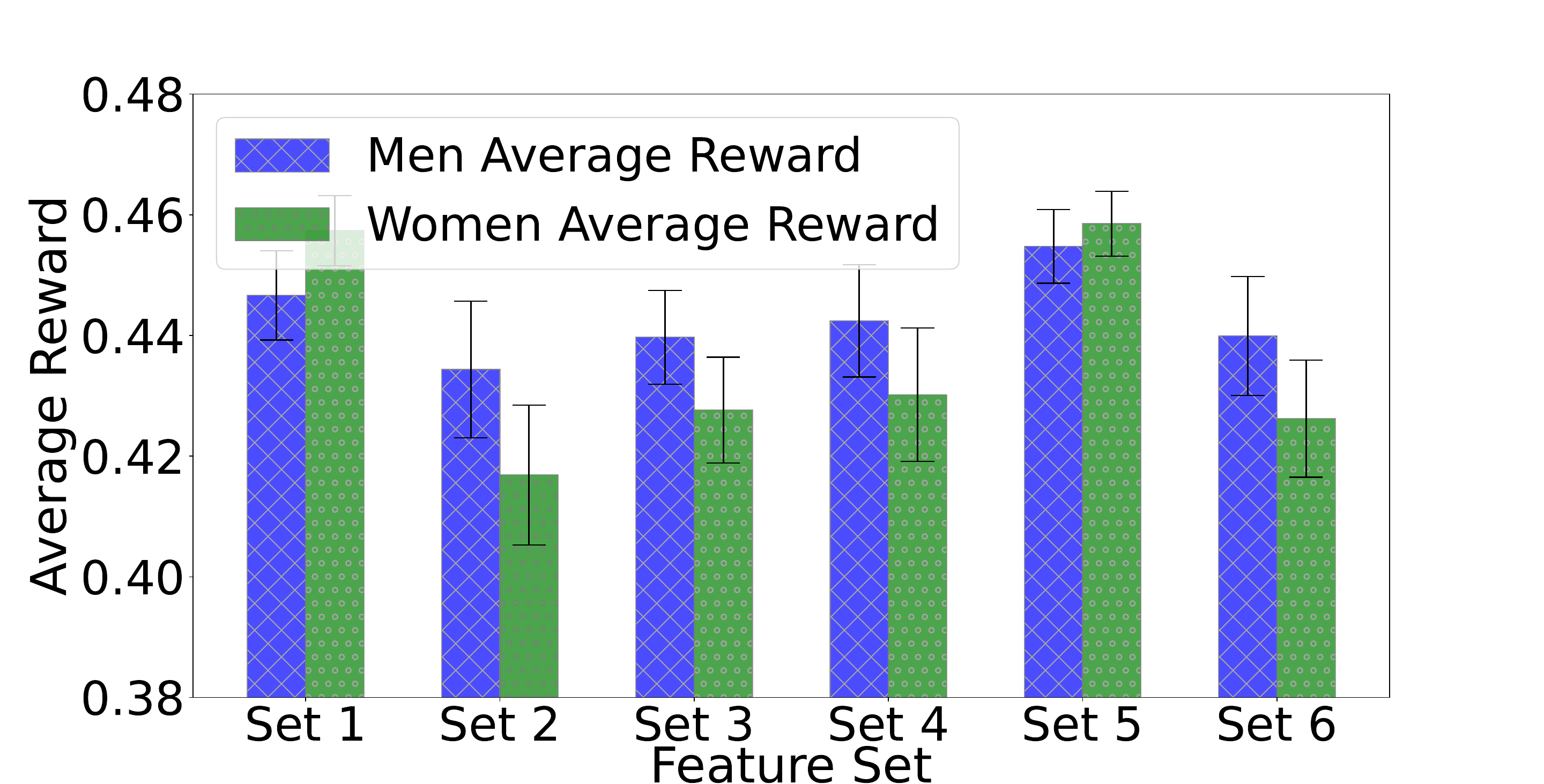}
    \captionof{figure}{Numerical means of the average reward over $100$ runs}
    \label{fig:Reward}
\end{minipage}
\hfill
\begin{minipage}{0.497\textwidth}
    \includegraphics[width=1.1\textwidth]{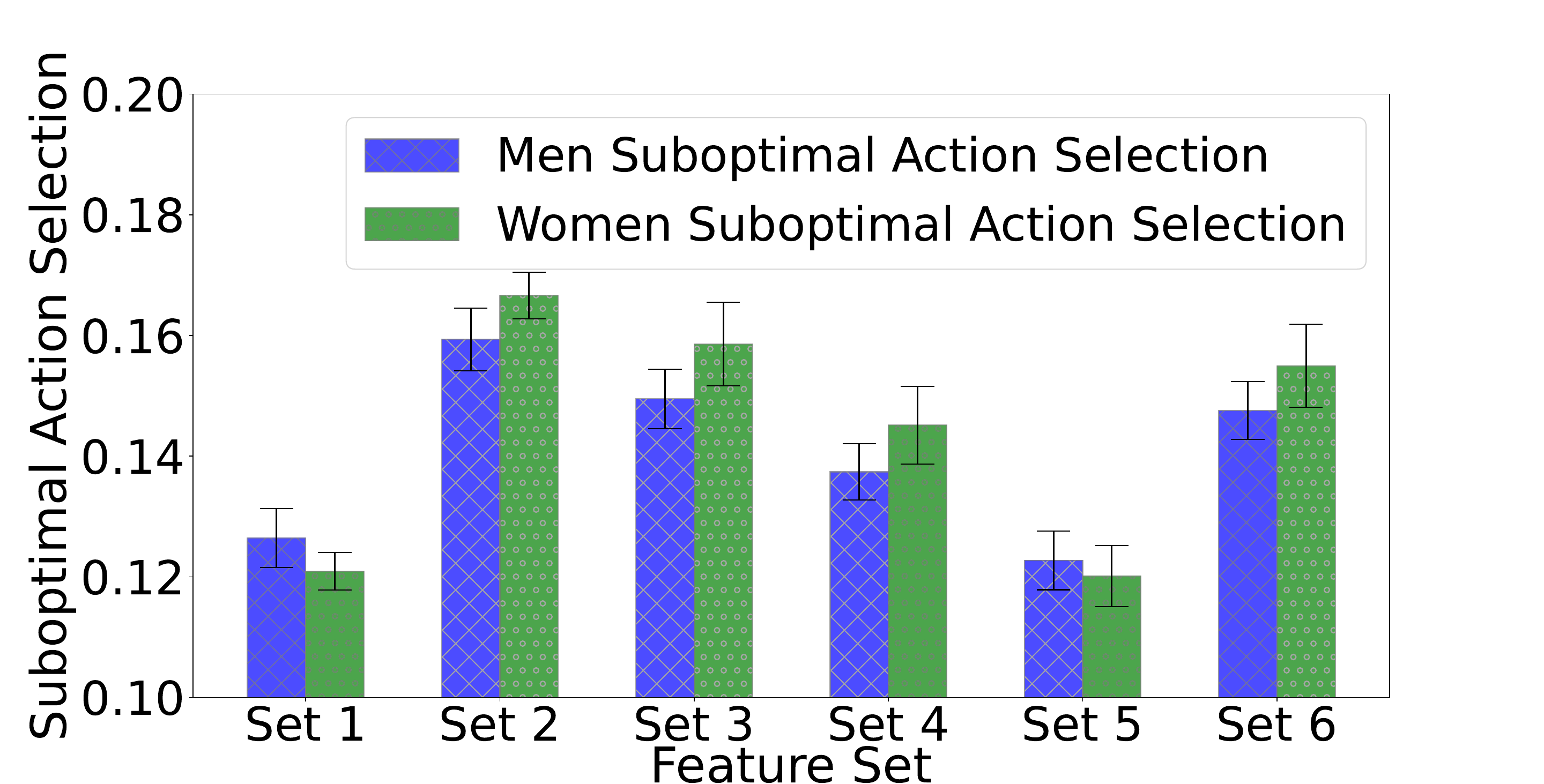}
    \caption{Numerical means of the fraction of suboptimal action selection over $100$ runs}
    \label{fig:SuboptimalAction1}
\end{minipage}
\end{figure}
\begin{table}[H]
  \centering 
  \caption{Statistical summary of the received average reward} %Hypothesis \textcolor{violet}{H1} is that the performance of the RL algorithm for women is better than the performance for men. Hypothesis \hyperlink{H2}{\textcolor{teal}{H2}} is that the performance of the RL algorithm for women is worse than the performance for men. Hypothesis \hyperlink{H3}{\textcolor{purple}{H3}} is that the performance of the RL algorithm for women is unequal to the performance for men.} 
  \begin{tabular}{c  >{\raggedleft\arraybackslash}p{1cm}  >{\raggedleft\arraybackslash}p{1cm}  >{\raggedleft\arraybackslash}p{1cm}  >{\raggedleft\arraybackslash}p{1cm}  c c c c}
    \toprule
    \multirow{2}{*}{\textbf{Feature Set}} & \multicolumn{2}{c}{\textbf{Men}} & \multicolumn{2}{c}{\textbf{Women}} & \multirow{2}{*}{\textbf{Hypothesis}} &  \multirow{2}{*}{\textbf{p-value}} &  \multirow{2}{*}{\textbf{Cohen's d}} \\
        \cline{2-3} \cline{4-5}
         & \textbf{Mean} & \textbf{Std} & \textbf{Mean} & \textbf{Std} \\
    \midrule
   Feature Set 1 & \underline{0.446} & 0.037 & \underline{0.457} & 0.029 &  \hyperlink{H1}{\textcolor{violet}{H1}}  & 0.012 & 0.322 \\ 
    Feature Set 2 & \underline{0.434} & 0.057 & \underline{0.416} & 0.058 & \hyperlink{H2}{\textcolor{teal}{H2}} & 0.017 & 0.303 \\ 
    Feature Set 3 & \underline{0.439} & 0.039 & \underline{0.427} & 0.044 & \hyperlink{H2}{\textcolor{teal}{H2}} & 0.021 & 0.288 \\
    Feature Set 4 & \underline{0.442} & 0.046 & \underline{0.430} & 0.055 & \hyperlink{H2}{\textcolor{teal}{H2}} & 0.047 & 0.237 \\
    Feature Set 5 & \underline{0.455} & 0.031 & \underline{0.458} & 0.027 & \hyperlink{H3}{\textcolor{purple}{H3}} & 0.362 & NA\\
    Feature Set 6 & \underline{0.439} & 0.050 & \underline{0.426} & 0.048 & \hyperlink{H2}{\textcolor{teal}{H2}} & 0.025 & 0.278\\ 
    \bottomrule
  \end{tabular}
  \label{tab:RewardSummary} 
\end{table}
\begin{table}[H]
  \centering 
  \caption{Statistical summary of the fraction of suboptimal action selection}% Hypothesis \textcolor{blue}{\hyperlink{H1}{\textcolor{violet}{H1}}} is that the performance of the RL algorithm for women is better than the performance for men. Hypothesis \hyperlink{H2}{\textcolor{teal}{H2}} is that the performance of the RL algorithm for women is worse than the performance for men. Hypothesis \hyperlink{H3}{\textcolor{purple}{H3}} is that the performance of the RL algorithm for women is unequal to the performance for men.} 
  \begin{tabular}{c  >{\raggedleft\arraybackslash}p{1cm}  >{\raggedleft\arraybackslash}p{1cm}  >{\raggedleft\arraybackslash}p{1cm}  >{\raggedleft\arraybackslash}p{1cm}  c c c c}
    \toprule
    \multirow{2}{*}{\textbf{Feature Set}} & \multicolumn{2}{c}{\textbf{Men}} & \multicolumn{2}{c}{\textbf{Women}} & \multirow{2}{*}{\textbf{Hypothesis}} &  \multirow{2}{*}{\textbf{p-value}} &  \multirow{2}{*}{\textbf{Cohen's d}} \\
        \cline{2-3} \cline{4-5}
         & \textbf{Mean} & \textbf{Std} & \textbf{Mean} & \textbf{Std} \\
    \midrule
   Feature Set 1 & \underline{0.126} & 0.023 & \underline{0.120} & 0.029 & \hyperlink{H1}{\textcolor{violet}{H1}} & 0.046 & 0.239\\ 
     Feature Set 2  & \underline{0.159} & 0.026 & \underline{0.166} & 0.019 & \hyperlink{H2}{\textcolor{teal}{H2}} & 0.014 & 0.313 \\ 
    Feature Set 3 & \underline{0.149} & 0.025 & \underline{0.158} & 0.035 & \hyperlink{H2}{\textcolor{teal}{H2}} & 0.018 & 0.299 \\
    Feature Set 4 & \underline{0.137} & 0.023 & \underline{0.145} & 0.032 & \hyperlink{H2}{\textcolor{teal}{H2}} & 0.027 & 0.273\\
    %Feature Set 5 & 0.179 & 0.021 & 0.174 & 0.011 & \hyperlink{H1}{\textcolor{violet}{H1}} & 0.031 & 0.265 \\
    %\rowcolor{gray!30} Feature Set 6 & 0.324 & 0.026 & 0.333 & 0.039 & \hyperlink{H2}{\textcolor{teal}{H2}} & 0.024 & 0.279  \\
    %Feature Set 7 & 0.129 & 0.026 & 0.120 & 0.031 & \hyperlink{H1}{\textcolor{violet}{H1}} & 0.018 & 0.297 \\
    %Feature Set 8 & 0.111 & 0.015 & 0.106 & 0.022 & \hyperlink{H1}{\textcolor{violet}{H1}} & 0.040 & 0.248 \\
    Feature Set 5 & \underline{0.122} & 0.024 & \underline{0.120} & 0.015 & \hyperlink{H3}{\textcolor{purple}{H3}} & 0.323 & NA\\
    Feature Set 6 & \underline{0.147} & 0.024 & \underline{0.155} & 0.034 & \hyperlink{H2}{\textcolor{teal}{H2}} & 0.040 & 0.248 \\ 
    \bottomrule
  \end{tabular}
  \label{tab:SuboptimalSumamry} 
\end{table}

%\caption*{The error bars represent the $95\%$ confidence intervals. In these bar plots, the Y-axis starts from non-zero values ($0.38$ and $0.1$) to emphasize the relative differences between the bars. Exact values are provided in Table \ref{tab:RewardSummary} and Table \ref{tab:SuboptimalSumamry} for reference. To ensure transparency, we provide alternative versions of these figures where the Y-axis starts from $0$ in the Appendix.}

%In Table \ref{tab:RewardSummary} and \ref{tab:SuboptimalSumamry}, we report the results along with respective standard deviations. 
To see how these sets affect gender fairness, we tested the following hypotheses\thinspace:
\begin{itemize}
    \item \hypertarget{H1}{\textcolor{violet}{H1}}  \textemdash \  the performance of the RL algorithm for women is \textit{better} than its performance for men.
    \item \hypertarget{H2}{\textcolor{teal}{H2}} \textemdash \ the performance of the RL algorithm for women is \textit{worse} than its performance for men.
    \item \hypertarget{H3}{\textcolor{purple}{H3}} \textemdash \ the performance of the RL algorithm for women is \textit{unequal} to its performance for men.
\end{itemize}

We used LinUCB with feature sets $1$ to $6$ on $50000$ patient interactions. Each patient interaction corresponds to a single time step in the results. We used the average reward and fraction of suboptimal action selection as performance measures. In this context, a suboptimal action is any action other than the action that has the highest expected reward for the considered patient interaction. All the results presented in this article are averaged over $100$ independent runs.
%See Figure \ref{fig:Reward}, Figure \ref{fig:SuboptimalAction1}, Table \ref{tab:RewardSummary} and Table \ref{tab:SuboptimalSumamry}. 

In Figure \ref{fig:Reward} and Table \ref{tab:RewardSummary}, we show the average reward obtained for men and women respectively.
 Similarly, we report the fraction of suboptimal action selection in Figure \ref{fig:SuboptimalAction1} and Table \ref{tab:SuboptimalSumamry}. In Figures \ref{fig:Reward} and \ref{fig:SuboptimalAction1}, 
the error bars represent $95\%$ confidence intervals to showcase the variability of the results. %In these figures, the Y-axis starts from non-zero values ($0.38$ and $0.1$) to emphasize the relative differences between the bars. 
In Figure \ref{fig:Reward}, the Y-axis starts at $0.38$, while in Figure \ref{fig:SuboptimalAction1} the Y-axis starts at $0.10$. The choice of using these non-zero values is made to emphasize the relative differences between the bars. Exact values for the average reward and fraction of suboptimal action selection are provided in Table \ref{tab:RewardSummary} and Table \ref{tab:SuboptimalSumamry} for reference. To ensure transparency, we provide alternative versions of these figures where the Y-axes start from $0$ in Appendix \ref{app:Results}.

In Table $\ref{tab:RewardSummary}$, the p-value of \textcolor{violet}{H1} for feature set $1$ is 0.012, which is below the significance level of $0.05$. Therefore, based on this conventional threshold, we may reject the null hypothesis in favor of hypothesis \textcolor{violet}{H1} for feature set 1. For feature sets $2$, $3$, $4$ and $6$, the respective p-values of \textcolor{teal}{H2} are $0.017$, $0.021$, $0.047$ and $0.025$.  Therefore,  based on the significance level of $0.05$, we may reject the null hypothesis in favor of hypothesis \textcolor{teal}{H2} for feature sets $2$, $3$, $4$ and $6$. Crucially, for feature set $5$, the p-value of \textcolor{purple}{H3} is $0.362$, indicating that there is insufficient evidence %against \textcolor{purple}{H3} at the conventional significance level of $0.05$. 
to reject the null hypothesis in favor of \textcolor{purple}{H3} at the conventional significance level of $0.05$. %Therefore, we fail to reject \textcolor{purple}{H3} for feature set $5$. 

Even for the performance measure of the fraction of suboptimal action selection shown in Table \ref{tab:SuboptimalSumamry}, we see similar trends as for average reward. For feature set $1$, we may reject the null hypothesis in favor of hypothesis \textcolor{violet}{H1} (p-value $0.046$). For feature sets $2$, $3$, $4$ and $6$, we may reject the null hypothesis in favor of hypothesis \textcolor{teal}{H2} (p-values $0.014, 0.018, 0.027$ and $0.040$ respectively). For feature set $5$, there is insufficient evidence to reject the null hypothesis in favor of \textcolor{purple}{H3} (p-value $0.323$).

We also report Cohen's d values which are commonly used in healthcare research to quantify the effect size of differences between groups, aiding in the evaluation of clinical significance \citep{Sullivan2012}.

The results suggest that for both the performance metrics, the gender fairness provided by the algorithm is affected by the features included in the context information. 

\subsection{Results\thinspace: Dynamic Selection of Optimal Feature Set} 
Here our goal is to verify the following\thinspace:
\begin{itemize}
    \item Does \PropAlg \ learn to select the optimal feature set? -- Learning to select the optimal feature set is important as the included features can affect gender fairness as well as the algorithm's utility.  
    \item Can \PropAlg \ make use of clinicians' domain knowledge to improve its performance and avoid cold start? -- Cold start here refers to the phenomenon of subpar initial performance for reinforcement learning solutions. See Section \ref{sec:OurCont}  and \ref{sec:GenInsights} for more details. 
\end{itemize}
The identity of the optimal feature set is determined by the performance criterion such as\thinspace:
\begin{itemize}
    \item optimal feature set = feature set with the highest adherence to some utility criterion (e.g., highest average reward);    
    \item optimal feature set = feature set with the highest adherence to some fairness criterion (e.g., the lowest difference between average rewards for men and women);
    \item weighted combination of the utility criterion and the fairness criterion. 
\end{itemize}
\begin{proposition}
\label{prop:PerformanceCriterion}
In our experiments, the performance criterion for feature sets is a weighted combination of the utility criterion and the fairness criterion with equal weights for both. \end{proposition}

It should be emphasized that \PropAlg \ is agnostic toward the choice of the performance criterion, and it can be used with any arbitrary performance criterion. 

We used \PropAlg \ along with Thompson Sampling (given in Appendix  \ref{app:TS}) as Policy1 to select the feature sets and LinUCB (given in Appendix \ref{app:LinUCB}) as Policy2 to select the pain care recommendations. 
As noted earlier, Thompson Sampling allows us to make use of clinicians' domain knowledge about the suitability of feature sets.  If the healthcare experts' beliefs about the suitability of feature sets approximate reality well, then \PropAlg \ would be able to converge on the optimal feature set sooner. On the other hand, even in the case that healthcare experts' beliefs are imprecise/incorrect, \PropAlg \ would still be able to converge on the optimal feature set by utilizing the observations in the form of \{selected feature set, corresponding reward\} to counteract the provided incorrect information. 
%This can be seen in Figure \ref{fig:3} and Figure \ref{fig:4}. 

\begin{figure}[!t]
\centering
\begin{minipage}{.5\textwidth}
  \centering
  \includegraphics[width=1.05\linewidth]{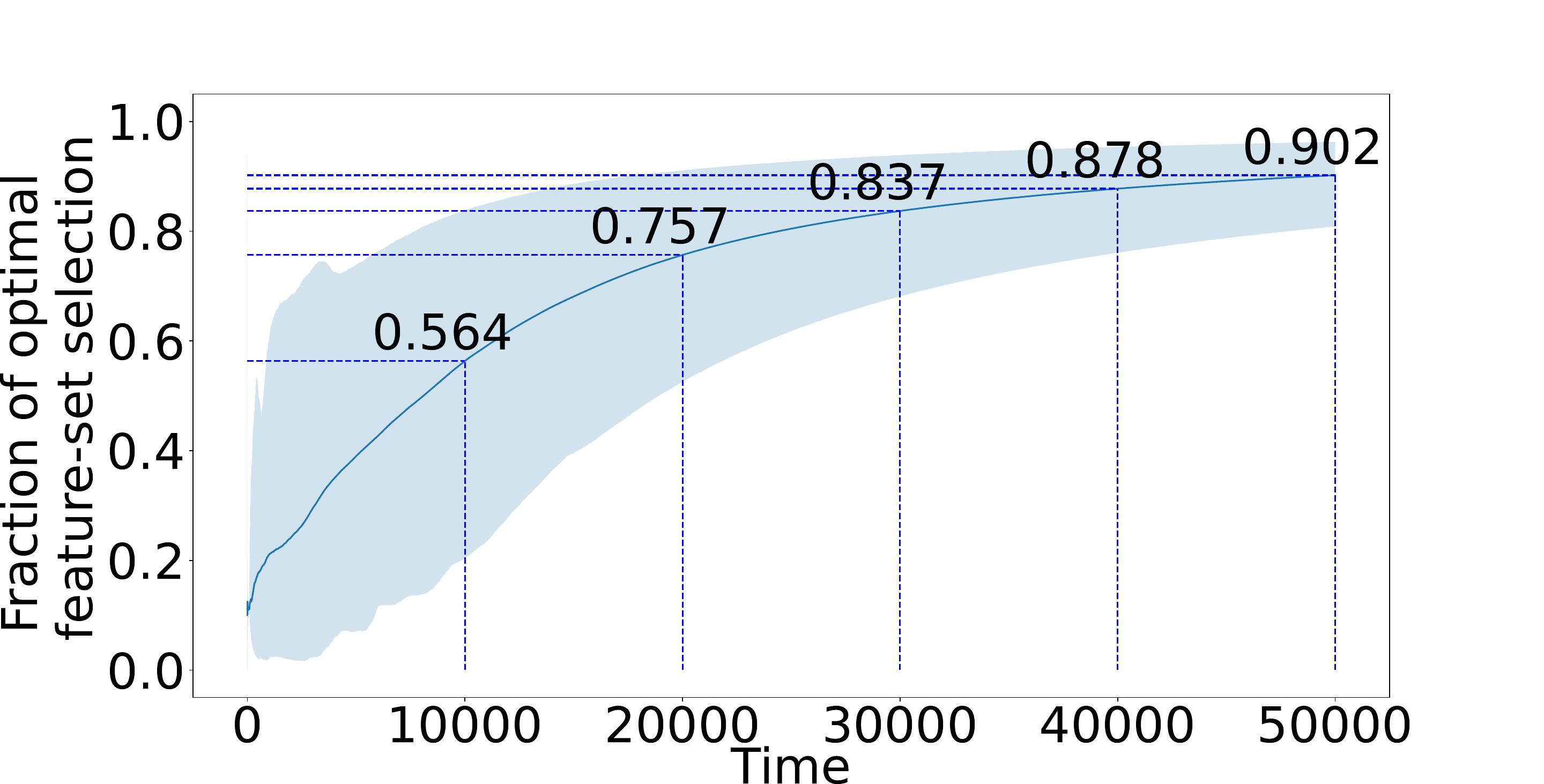}
  %\captionof{figure}{Fraction of Optimal Feature Set Selection from $T=1$ to $50000$. The shaded areas represent the variability in the results, extending from the minimum to the maximum value.}
\end{minipage}%
\begin{minipage}{.5\textwidth}
  \centering
  \includegraphics[width=1.05\linewidth]{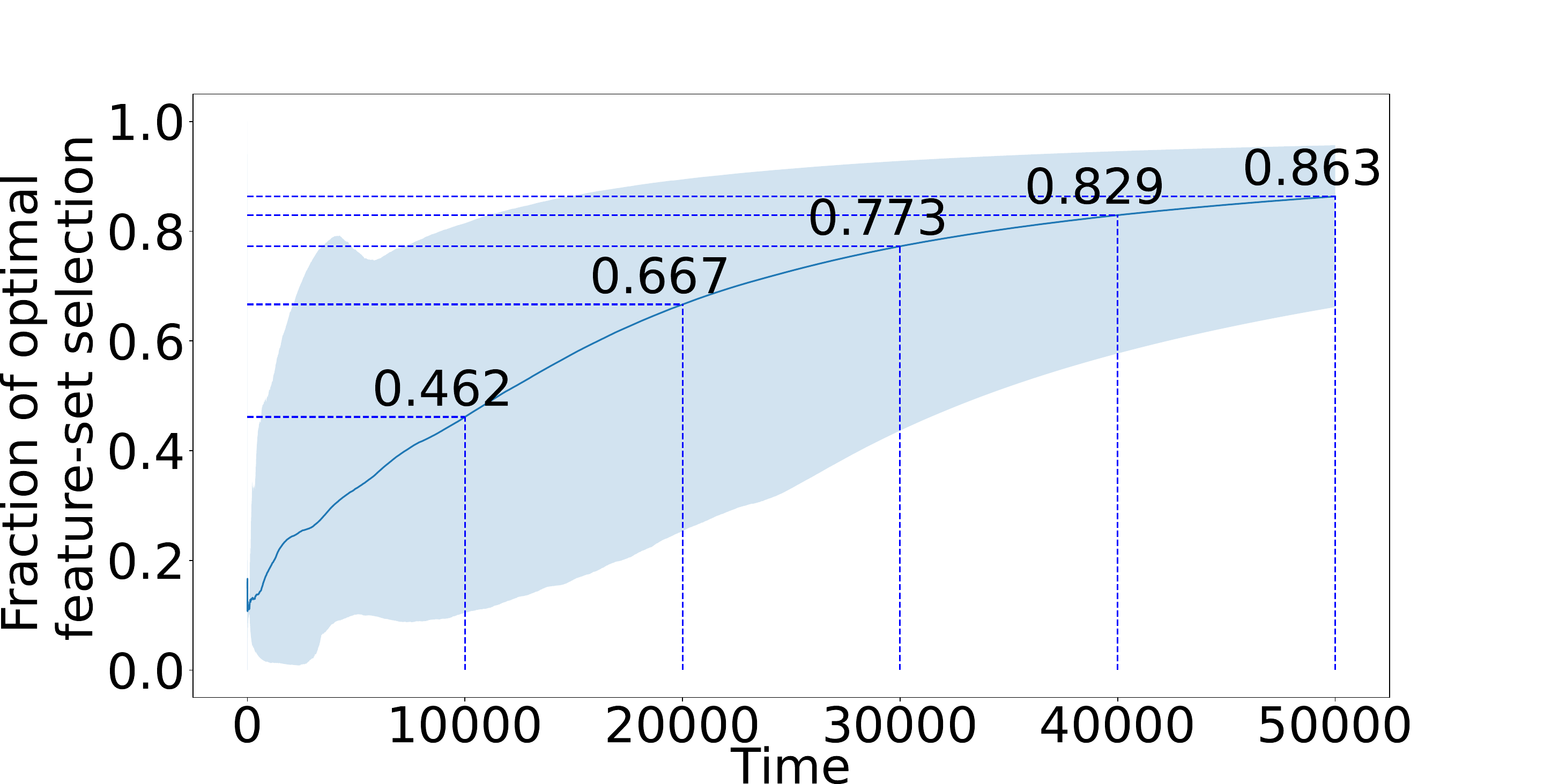}
  %\captionof{figure}{Fraction of Optimal Feature Set Selection from $T=1$ to $50000$. The shaded areas represent the variability in the results, extending from the minimum to the maximum value.}
\end{minipage}
\caption{Cumulative fraction of optimal feature set selection from $T=1$ to $50000$. The shaded areas represent the variability in the results, extending from the minimum to the maximum value over $100$ runs. The results on the left are obtained using a prior distribution that more closely approximates the suitability of each feature set for the task at hand.}
\label{fig:3}
%\end{figure} 
%\begin{figure}
\centering
\begin{minipage}{.5\textwidth}
  \centering
  \includegraphics[width=1.05\linewidth]{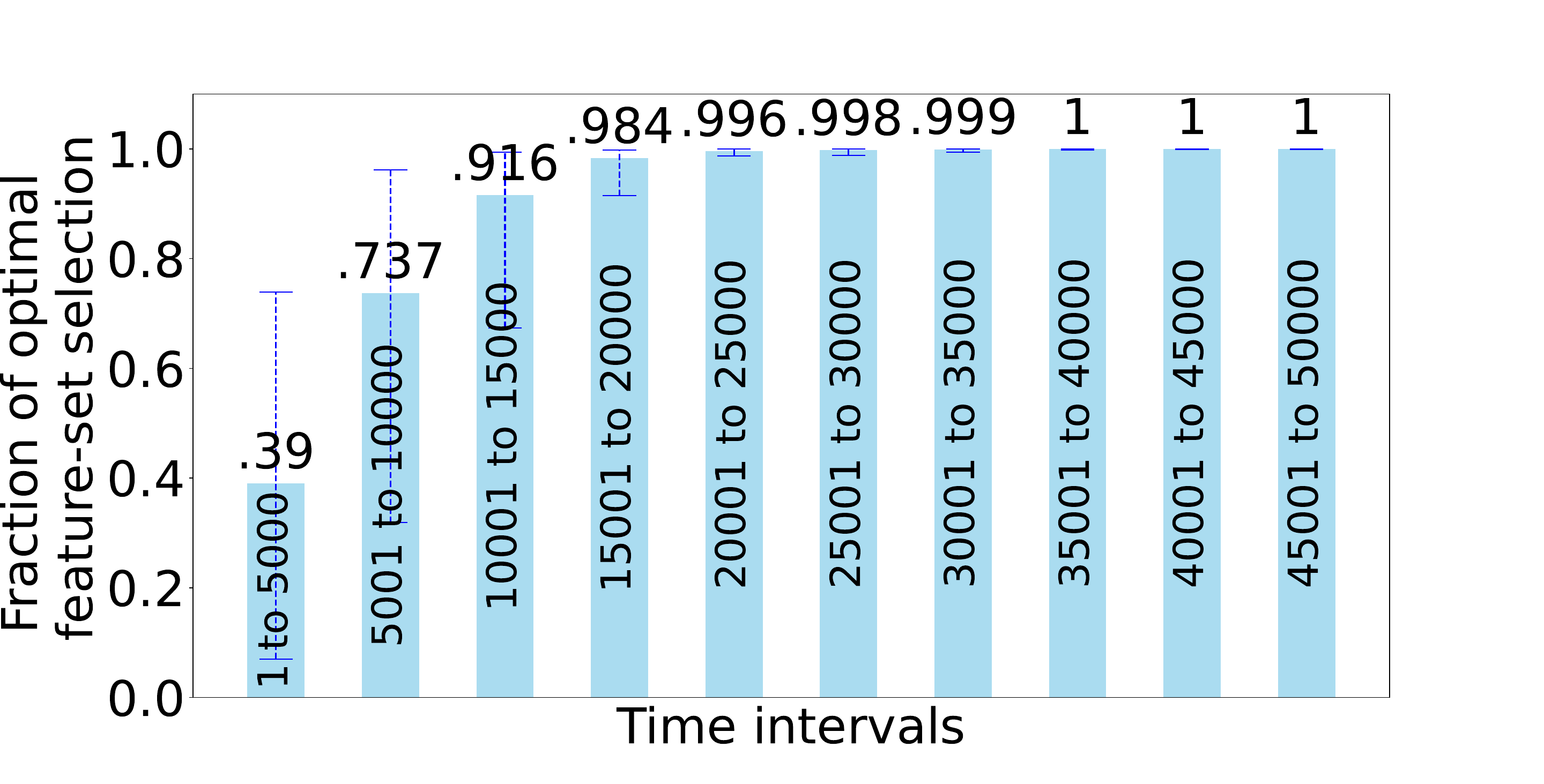}
  %\captionof{figure}{Fraction of Optimal Feature Set Selection from $T=1$ to $50000$. The shaded areas represent the variability in the results, extending from the minimum to the maximum value.}
  \label{fig:T3000AvgRewardFeatureSet}
\end{minipage}%
\begin{minipage}{.5\textwidth}
  \centering
  \includegraphics[width=1.05\linewidth]{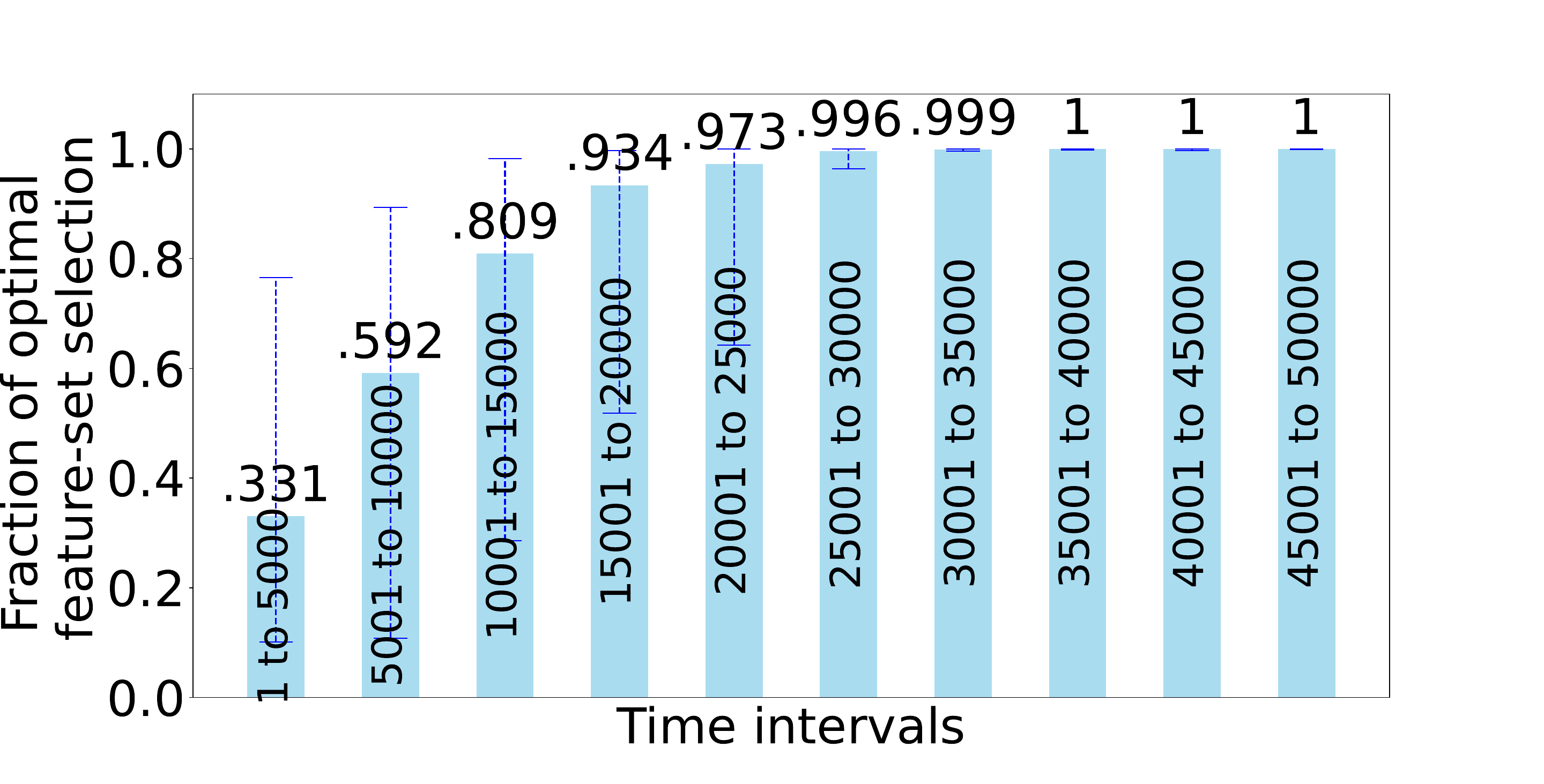}
  %\captionof{figure}{Fraction of Optimal Feature Set Selection from $T=1$ to $50000$. The shaded areas represent the variability in the results, extending from the minimum to the maximum value.}
  \label{fig:T5000AvgRewardFeatureSet}
\end{minipage}
\caption{Fraction of optimal feature set selection in the $10$ intervals of $5000$ up to $T=50000$. The error bars show the variability in the results, extending from the minimum to the maximum value over $100$ runs. The results on the left are obtained using a prior distribution that more closely approximates the suitability of each feature set for the task at hand.}
\label{fig:4}
\end{figure} 
In Figure \ref{fig:3}, we plot time on the X-axis and the cumulative fraction of optimal feature set selection on the Y-axis. To see how to interpret Figure \ref{fig:3}, consider a point corresponding to the X value of $10000$. In this context, the corresponding Y value is $0.564$. This value indicates that, from time step $1$ to $10000$, the algorithm opted for the optimal feature set approximately $56.4\%$ of the time (averaged over $100$ runs). %Similarly the point $(20000, 0.757)$ indicates that from time step $1$ to $20000$, the algorithm opted for the optimal feature set approximately $75.7\%$ of the time (averaged over $100$ runs).  

In Figure \ref{fig:4}, we plot time intervals on the X-axis and the fraction of optimal feature set selection on the Y-axis. To see how to interpret Figure \ref{fig:4}, consider the second from the left bar corresponding to the time interval $5001$ to $10000$. The height of this bar is 0.737 which indicates that from time step $5001$ to $10000$, the algorithm selected the optimal feature set approximately $73.7\%$ of the time (averaged over $100$ runs).

From the results given in Figure \ref{fig:3} and Figure \ref{fig:4}, it can be seen that \PropAlg \ learns to select the optimal feature set. In  Figure \ref{fig:3}, the cumulative fraction of optimal feature set selection increases with time. In Figure \ref{fig:4}, the fraction of the optimal feature set selection in the considered time intervals increases rapidly until it reaches $1$, indicating that beyond a certain time step, the algorithm always makes pain care recommendations using the optimal feature set.

%The results given on the left of Figure \ref{fig:3} and Figure \ref{fig:4} are obtained using prior distributions constructed using clinicians' domain knowledge. 
The results on the left of Figure \ref{fig:3} and Figure \ref{fig:4} are obtained by setting the prior distribution for all the actions as $\textrm{Beta}(1,2)$ which encodes the belief that the expected reward from each feature set would be  $1/(1+2) = 0.333$ which is not too far from reality as seen in Table \ref{tab:RewardSummary}.  The results on the right are obtained by setting the prior distribution for all the actions as $\textrm{Beta}(1,5)$ which encodes the belief that the expected reward from each feature set would be  $1/(1+5) = 0.167$. 

Comparing the results on the left with the respective results on the right in Figure \ref{fig:3} and Figure \ref{fig:4}, it can be seen that \PropAlg \ is indeed able to identify the optimal feature set sooner when the prior distribution approximates reality better. For example, in Figure \ref{fig:4}, see the bars for the time interval `$5001$ to $10000$' on the left and the right.  On the left, this fraction is $0.737$, and on the right, this fraction is $0.592$. That is \PropAlg \ selected the optimal feature set much more frequently when the clinicians' domain knowledge about the feature sets approximates reality better. However, even in the other case, when these beliefs are imprecise, \PropAlg \ is still able to converge on the optimal feature set as seen in the bars for time intervals `$35001$ to $40000$', `$40001$ to $45000$' and `$45001$ to $50000$' on the right in Figure \ref{fig:4}. The effect of healthcare experts' beliefs on the RL algorithm's performance is also seen in Figure \ref{fig:3}. On the left, from time step $1$ to $10000$, the algorithm selected the optimal feature set $56.4\%$ of the time, while on the right, for the same time period, the algorithm selected the optimal feature set $46.2\%$ of the time. This indicates that the algorithm is able to identify the optimal feature set sooner when the prior distribution constructed using clinicians' domain knowledge resembles reality. This leads to more frequent selection of the optimal feature set from the outset, resulting in higher performance based on the given criterion (in this case, a weighted combination of utility and gender fairness with equal weights for both, as given in Proposition \ref{prop:PerformanceCriterion}) even during initial time steps.
 Thus, \PropAlg \ demonstrates the capability to mitigate cold start to some degree by enhancing its initial performance through leveraging clinicians' domain expertise.

\section{Discussion} 
%-{\color{red} \emph{This is probably the most important section of your paper!  This  is where you tell us how your work advances our understanding of  machine learning and healthcare.}  Discuss both technical and clinical implications, as appropriate\cite{xyz19}.}
%{\color{red}
%Explain when your approach may not apply, or things you could not
%check.  \emph{Discussing limitations is essential.  Both ACs and
%  reviewers have been advised to be skeptical of any work that does
%  not consider limitations.} }  

\paragraph{Machine learning perspective.} We demonstrate how algorithms can display biased outcomes with respect to gender on a real-world dataset. Our approach showcases that information considered in the decision-making process can affect gender fairness provided by the RL algorithm. Clinicians may find it valuable to understand which specific details of the inclusion or exclusion of patient information can result in biased outcomes. This understanding can help them select the most appropriate information to generate pain care recommendations that are both highly useful and aligned with fairness principles. Our proposed RL solution, called \PropAlg, can automate this selection process and dynamically learn to use the optimal information set that achieves the given objective. We also showcased a potential solution to alleviate cold start by enhancing the initial performance of our solution by leveraging clinicians' domain expertise. Moreover, this approach offers a way for clinicians to maintain their advisory role in ML-driven pain care recommendation systems, addressing concerns about the loss of this role as raised in \cite{Grote2022}.
A future direction of this work could be incorporating a deep learning approach to address disparities in pain, as done for racial disparities by \cite{Pierson2021}. However, deep learning approaches are generally considered to be less interpretable/explainable compared to traditional ML approaches which might be a hindrance to the former's acceptability among clinicians and policymakers. 
%PG(8 April): John, I suggest adding the above two statements which refer to the paper from one of my previous comments. Is the bit about explainability being a  concern for acceptability among clinicians and policymakers correct? 

\paragraph{Limitations from ML perspective (and avenues for future work).}
\begin{itemize}
    %\item Results on only one dataset\thinspace: Conducting experiments with a wider range of datasets and exploring other protected attributes would enhance the general applicability of our findings. 
    \item Exploring the cause(s) behind demonstrated gender bias\thinspace: It would be illustrative to see whether this bias is an artifact of the underlying data or it is introduced/amplified by the RL algorithm.
    \item Other definitions of fairness\thinspace: The considered definition of fairness in this work is a group-based notion. Various other notions of fairness could be considered in this context \citep{barocas-hardt-narayanan, GajPechFAT2018, Grote2022}.
    \item Formulation of chronic pain care management\thinspace: Formulation as a Markov decision process instead of contextual bandits may more accurately represent the real-world dynamics of the problem %in which patient's previous ``states" and received recommendations affect their current ``state" 
    \citep{Steimle2017}.
    %\item Scalability of results -- 
    \item Unsuitable for non-stationary environment\thinspace: The efficacy of treatments or the identity of the optimal feature set can change over time due to physiological changes \citep{PMID:24547798}, psychosocial factors \citep{turk2018psychological} and environmental influences \citep{karos2019pain}. Such a non-stationary environment would require a change in our approach for the detection of these changes and the subsequent adjustment in the action selection if necessary, especially in scenarios where no prior information about non-stationarity is available \citep{pmlr-v99-auer19b}.
    \item Risk-oblivious recommendations\thinspace: Modifying the problem such that the RL algorithm can adaptively defer to a baseline policy or a human expert when RL-based decisions are deemed unsafe/unfair will enhance the suitability of this work for safety-critical and responsible healthcare applications \citep{joshi2023learningtodefer, 10.5555/3327345.3327513}. 
\end{itemize}

\paragraph{Clinical perspective.} Our study validates the concerns raised by researchers, policymakers, and clinicians that machine learning algorithms may contribute to disparities or unfairness in patient care or patient outcomes due to limitations in the learning dataset(s) or the design of the RL algorithm. Specifically, here we highlight the important ways in which the selection of features may influence not only overall system performance (as measured by the overall average reward) but also disparities or differences in performance (which might be deemed unfair) across groups of patients. The current study goes beyond empirically illustrating that such algorithmic limitations can exist. Although the feature selection methodology we propose can undoubtedly be improved, it has broader significance in that it refutes the tendency to see fairness issues in machine learning as inevitable flaws that should prevent adoption in health systems. Rather, we demonstrate that empirical patient data can be employed to optimize rewards and minimize disparities across groups in ways that may even address long-standing differences in treatment quality, access, and outcomes across groups of patients currently receiving care in health systems around the world \citep{Green2005-re}. 

\paragraph{Limitations from clinical perspective (and avenues for future work).}
A strength of this study is that it is based on data from a large peer-reviewed clinical trial of an RL-based healthcare application. While this is the largest study of its kind published as of 2022, it still only represents data collected from weekly interactions with $168$ patients with chronic pain recruited from two United States Department of Veterans Affairs (VA) health systems. VA patients are known to be unique in terms of their clinical and sociodemographic profiles, with somewhat higher proportions of patients who have a college education as well as a high prevalence of chronic psychiatric, substance use disorder, and medical comorbidities \citep{Chang2020}. In particular, women patients with chronic pain treated in VA health systems are substantially different from adult women in other healthcare systems and their experience may not be broadly generalizable with respect to key issues such as pain severity, engagement in treatment, and treatment response. 

While Figure \ref{fig:Reward} and Table \ref{tab:RewardSummary} clearly establish that statistically significant disparities in average RL rewards could be identified between men and women, the clinical significance of the difference between genders in the average reward (or other characteristics of the distributions in reward) is not clear. Moreover, our analyses have focused solely on one type of disparity resulting from the selection of RL features (and how this issue can be addressed algorithmically). In the current dataset as well as other applications of RL, a variety of additional disparities may exist, including disparities related to social determinants of health such as patients' racial identification, educational attainment, geographic access to care, or comorbidities.

We believe these analyses illustrate the important relationship between feature selection and possible fairness issues in RL applications in healthcare. Other key protocol design decisions could cause or ameliorate fairness issues and are not addressed by the current study. For example, the algorithm used in the current study employed a ``cost" coefficient to penalize action choices that require more clinician time and assist in RL decision-making.
%when two or more options had the same expected future reward. 
How that cost function is operationalized in the extreme case could completely drive the RL algorithm's decision-making. The best strategy for defining cost fairness or other performance characteristics is not well described. RL algorithms employ an explore/exploit approach to optimization, and that also is a dimension of system design that could affect fairness in ways not addressed by the current study. Finally, differential missingness in feature data (e.g., if women are less engaged than men, and report health and behavioral inputs less reliably) could have a profound differential effect on algorithm performance across subgroups. The current study has not addressed this key issue and we view it as a potential avenue for future research. 
% ACKNOWLEDGEMENTS ONLY GO IN THE CAMERA-READY, NOT THE SUBMISSION
% \acks{Many thanks to all collaborators and funders!}
%Do NOT change font size of references or modify the bibliography style
\bibliography{references}

\appendix
\section*{Appendices}
\addcontentsline{toc}{section}{Appendices}
\renewcommand{\thesubsection}{\Alph{subsection}}

\begin{comment}
\section{Details about the study conducted in \cite{Piette2022}}
\label{app:AICBT_Trial}
In a recently published randomized trial, \cite{Piette2022} evaluated the effectiveness of reinforcement learning as a strategy for increasing the efficiency of personalized  CBT for chronic pain. This approach automatically adjusts the type and duration of treatment sessions based on patients’ feedback regarding their pedometer-measured step counts and pain-related interference. They found that the RL algorithm was effectively able to personalize the intensity of sessions to participants and that on average patient-reported outcomes improved as the program gained experience through interactions \cite{PIETTE2022100064}. 
Compared to standard telephone CBT delivered by a therapist, the RL-supported model resulted in non-inferior improvements in pain-related disability at three months (primary outcome), while using less than half the therapist time \cite{PIETTE2022100064}.  At six months, significantly more patients randomized to the RL arm had clinically meaningful improvements in pain-related disability than patients receiving standard care ($37\%$ versus $19\%$, $p<.01$) \citep{Piette2022}. Also, a greater proportion of patients receiving RL-supported CBT had clinically meaningful 6-month improvements in pain intensity ($29\%$ versus $17\%, p=.03$).
\end{comment}

\section{Details about Features}
\label{app:Features}
\begin{table}[H]
  \centering
  \caption{Features taken into account in making treatment recommendations}
  \label{tab:features}
  \begin{tabular}{p{0.3\linewidth} p{0.7\linewidth}}
    \hline
    \hspace{1.5cm}\textbf{Feature} & \hspace{3.8cm}\textbf{How measured?} \\
    \hline
    Pedometer-measured steps\textsuperscript{(a)} & Proportion of days in the current week in which the patient met the goal of walking 10\% more than the average number of steps in the most recent prior week with available data. \\
    \hline
    Change in average pain intensity\textsuperscript{(b)} & The difference between the average pain intensity for the current week and the most recent prior week with available data. Change across weeks, measured between -100\% and +100\%, was normalized to a range of 0-1. \\
    \hline
    CBT skill practice this week\textsuperscript{(c)} & The average score across reporting days, measured on a scale of 0-10 and normalized between 0-1. \\
    \hline
    Sleep quality this week\textsuperscript{(d)} & The average score across reporting days, measured on a scale of 0-10 and normalized between 0-1. \\
    \hline
    Sleep duration this week\textsuperscript{(e)} & The average score across reporting days, measured on a scale of 0-22 and normalized between 0-1. \\
    \hline
    Pain-related interference Q1\textsuperscript{(f)} & The average across reporting days, measured on a scale of 0-10 and normalized between 0-1. \\
    \hline
    Pain-related interference Q2\textsuperscript{(g)} & The average across reporting days, measured on a scale of 0-10 and normalized between 0-1. \\
    \hline
    Session number & An indicator of whether the current session is the 1st, 2nd, \dots, 10th, normalized between 0-1. \\
    \hline
    Summary of rewards associated with prior IVR session recommendations & A weighted average of the rewards that AI-CBT-CP received in prior sessions (more weight on recent sessions) when recommending IVR sessions to the patient, normalized between 0-1. \\
    \hline
    Summary of rewards associated with prior recommendations for a 15-minute therapist session & A weighted average of the rewards that AI-CBT-CP received in prior sessions (more weight on recent sessions) when recommending 15-minute sessions to the patient, normalized between 0-1. \\
    \hline
    Summary of rewards associated with prior recommendations for a 45-minute therapist session & A weighted average of the rewards that AI-CBT-CP received in prior sessions (more weight on recent sessions) when recommending 45-minute sessions to the patient, normalized between 0-1. \\
    \hline
  \end{tabular}
\end{table}

\begin{enumerate}[label=(\alph*)]
    \item “How many steps did you take today?”
    \item “Rate your average level of pain today, using a 0 to 10 scale with 0 meaning no pain and 10 meaning the worst pain imaginable.”
    \item “Using a scale from 0 to 10, with 0 representing not at all accomplished and 10 representing completely accomplished, please enter how well you accomplished your [skill name] skill practice today”
    \item “Rate how refreshed or rested you felt after last night’s sleep, with 0 being not at all rested and 10 being extremely rested.”
    \item “How many hours were you asleep last night”
    \item “What number best describes how much pain has interfered with your enjoyment of life today, with 0 meaning pain does not interfere and 10 meaning pain completely interferes?”
    \item “What number best describes how much pain has interfered with your general activity today, with 0 meaning pain does not interfere and 10 meaning pain completely interferes?” 
\end{enumerate}

\section{LinUCB Algorithm}
\label{app:LinUCB}
\begin{algorithm}
    \caption{LinUCB}\label{alg:LinUCB}
    %\SetAlgoLined
    \SetKwInOut{Parameter}{Parameter}
    \Parameter{$\alpha \in \mathbb{R}$}
    \ForEach{$a \in \mathcal{A}$}{
        $\boldsymbol{A}_a \gets \boldsymbol{I}_d$ \tcp*[r]{d-dimensional Identity matrix}
        $\boldsymbol{b}_a \gets \boldsymbol{0}_{d \times 1}$ \tcp*[r]{d-dimensional zero vector}
    }
    \For{$t=1, 2, 3, \dots$}{
        Observe features $\boldsymbol{x}_t \in \mathbb{R}^d$.\\
        $\hat{\boldsymbol \theta}_a \gets \boldsymbol{A}_a^{-1} \boldsymbol{b}_a$.\\
        $p_{t,a} \gets  \hat{\boldsymbol \theta}_a \boldsymbol{x}_t  + \alpha \sqrt{ \boldsymbol{x}_t \boldsymbol{A}_a^{-1} \boldsymbol{x}_t}$.\\
        Choose arm $a_t = \arg\max_{a \in \mathcal{A}} p_{t,a}$ with ties broken arbitrarily, and observe a real-valued reward $r_t$. \\
        $\boldsymbol{A}_{a_t} \gets \boldsymbol{A}_{a_t} + \boldsymbol{x}_t \boldsymbol{x}_t^\top$.\\
        $\boldsymbol{b}_{a_t} \gets \boldsymbol{b}_{a_t} +  r_t \boldsymbol{x}_t$.
    }
\end{algorithm}

\FloatBarrier
\newpage
\section{Thompson Sampling}
\label{app:TS}
\begin{algorithm}
    \caption{Thompson Sampling} 
    \label{alg:TS}
    \SetAlgoLined
    \SetKwInOut{Input}{Input}
    \SetKwInOut{Output}{Output}
    \Input{Action set $\{1, 2, \dots, K\}$, parameters $\alpha_a$, $\beta_a$ for $a = 1, 2, \dots, K$}
    %\Output{Action to take $a(t)$}
    \BlankLine
    \For{time step $t=1,2, \dots$}{
         \underline{\textbf{Action Selection}}\; \\
        \For{each action $a = 1, 2, \dots, K$}{
            Sample $\theta_a(t)$ from the Beta$(\alpha_a, \beta_a)$ distribution\;
        }
        Select action $a(t) = \arg\max \theta_a(t)$ and receive reward $r(t)$\; \\
        \BlankLine
        \underline{\textbf{Policy Update}}\; \\
        Perform a Bernoulli trial with success probability $r(t)$ and observe sample $s(t)$\; \\
        \If{$s(t) = 1$}{
            $\alpha_{a(t)} = \alpha_{a(t)} + 1$\;
        }
        \Else{
            $\beta_{a(t)} = \beta_{a(t)} + 1$\;
        }
    }
\end{algorithm}
\FloatBarrier

\section{Preprocessing}
\label{app:Pre}
Directly utilizing a logged dataset for our evaluation experiments presents the challenge that rewards are solely recorded for actions chosen by the logging policy, which may diverge from those selected by the algorithm under evaluation. %To address this challenge, we devised a model to emulate patient interactions based on the interactions outlined in the dataset \cite{piette2022dataset2}. 
%In this model,
To address this challenge, we constructed a set of weights for each action, enabling the transformation of feature values into expected rewards. Using these weights, rewards are obtained for any action with some noise being added to the expected reward. We used zero-mean Gaussian noise with a standard deviation equal to the standard deviation observed in the recorded rewards for the corresponding action in the dataset \citep{piette2022dataset2}. To generate additional patient interactions, numerical means of feature values for men and women for each session and each cluster\footnote{Patients are divided into three clusters in the dataset \cite{piette2022dataset2}.} are calculated. 
Auxiliary feature values are obtained using the corresponding numerical means with an additive noise. We used zero-mean Gaussian noise with a standard deviation equal to the standard deviation observed in the recorded feature values. %While generating additional patient interactions, we ensured that the distribution of the total population across genders remains consistent with the dataset \cite{piette2022dataset2}. 
Overall, women constitute $12.5\%$ of the total number of patients, which is consistent with the dataset \cite{piette2022dataset2}. 

\section{Algorithmic Parameter Used in LinUCB}
\label{app:LinUCBParam}
The role of $\alpha$ in the algorithm design is to control the \textit{exploration}. 
%Here, exploration refers to the tendency to select actions that appear suboptimal based on the available information at the time of selection. 
Exploration is performed to acquire information on action profitability and lower values of $\alpha$ lead to less exploration. 
Inadequate exploration may cause the algorithm to overlook highly rewarding actions, while excessive exploration can impede the algorithm's long-term objective of maximizing cumulative reward by sacrificing immediate rewards. Thus, achieving an optimal balance of exploration is crucial. 
On our data, we ran LinUCB with $\alpha = \{0.1, 0.2, 0.3, \dots, 1\}$. 
%The value of $\alpha = 0.3$ performed the best in terms of the obtained utility. 
A trend resembling an inverted U-shape was observed in utility measured by average reward, with a global maximum at $\alpha=0.3$, and a decrease for both lower (i.e., $<0.3$) and higher (i.e., $>0.3$) values of $\alpha$. No discernible difference in gender fairness was observed across all tested values. Correspondingly, we use $\alpha = 0.3$ in all the results reported in this article.

\section{Auxiliary Results}
\label{app:Results}
%Some more details about those methods, so we can actually reproduce them.  After the blind review period, you could link to a repository for the code also.  \emph{MLHC values both rigorous evaluation as well  as reproduciblity.}

\begin{figure}[h]
    \centering
    \includegraphics[width=0.75\textwidth]{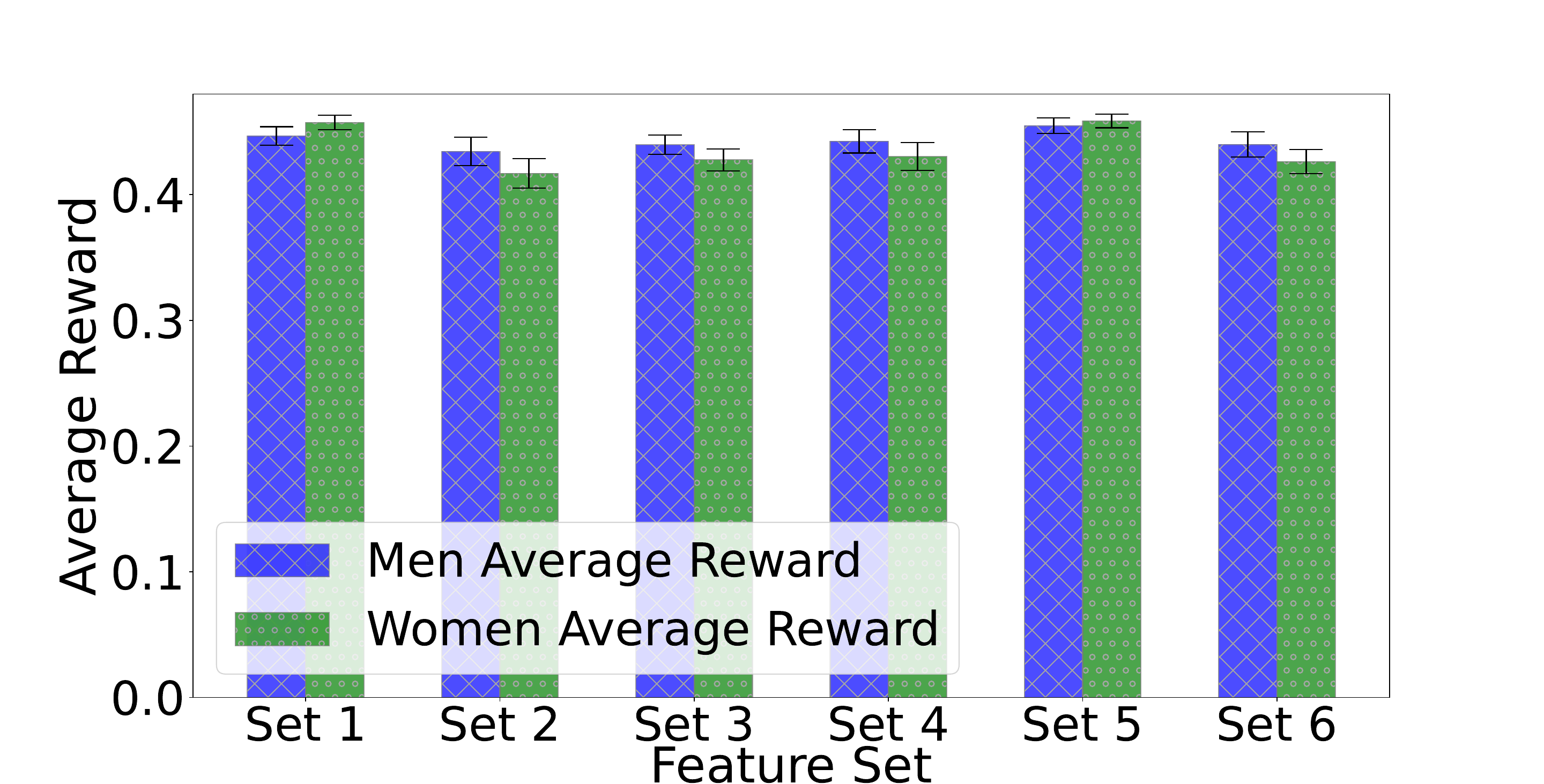}
    \caption{Numerical means of the average reward obtained for men and women over 100 runs. The error bars represent the $95\%$ confidence intervals.}
    \label{fig:RewardApp}
\end{figure}
\begin{figure}[h]
    \centering
    \includegraphics[width=0.75\textwidth]{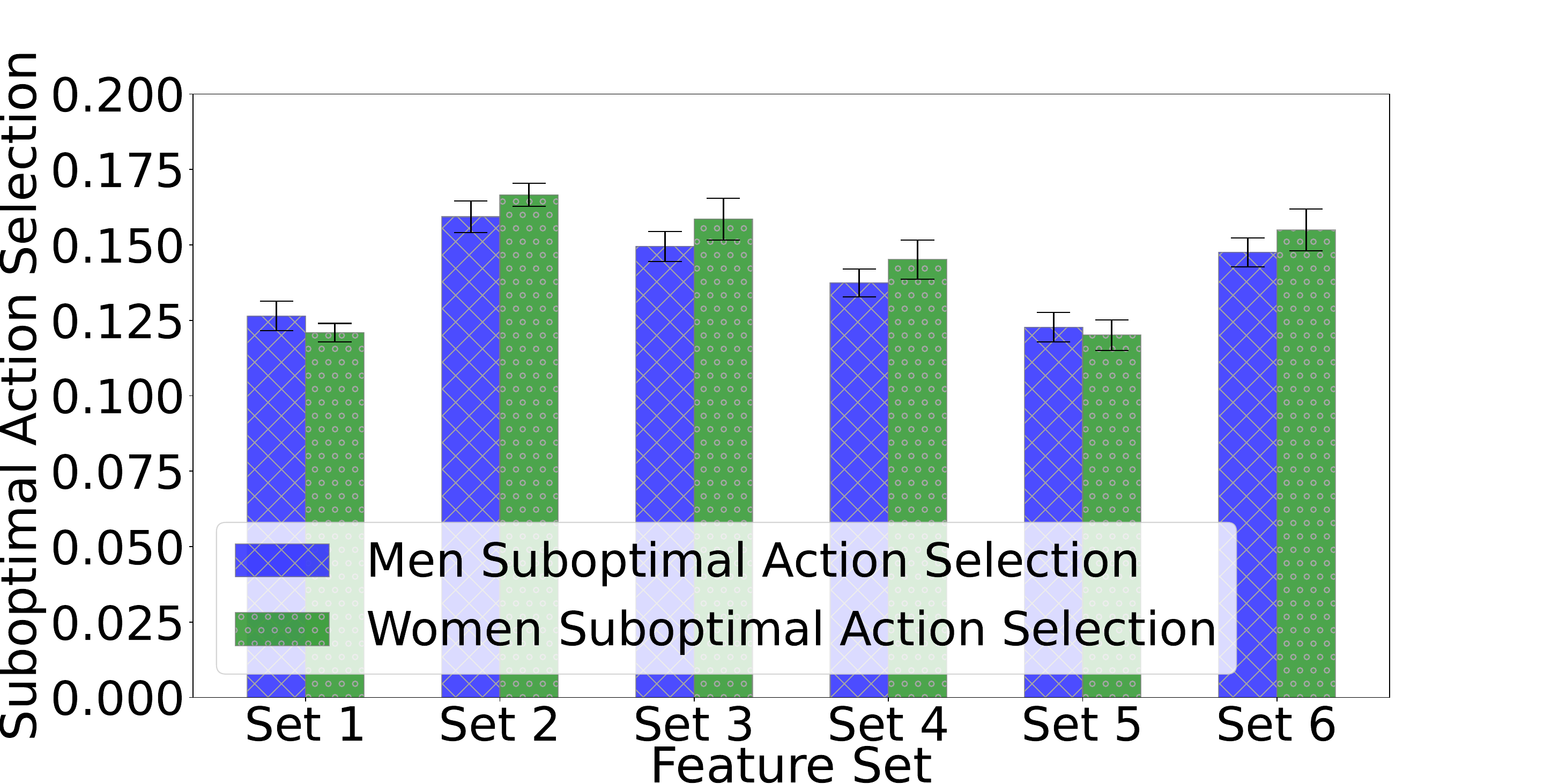}
    \caption{Numerical means of the fraction of suboptimal action selection for men and women over 100 runs. The error bars represent the $95\%$ confidence intervals.}
    \label{fig:SuboptimalAction1App}
\end{figure}
\end{document}